\documentclass{styles/svproc}

\usepackage{url}

\usepackage{array}
\usepackage{booktabs}
\usepackage{framed,multirow}
\usepackage{graphicx}
\usepackage{epsfig}
\usepackage{amsmath}
\usepackage{amssymb}
\usepackage{enumitem}

\usepackage{bm}
\usepackage{upgreek}
\usepackage{algorithmic}
\usepackage[vlined,ruled]{algorithm2e}
\usepackage{soul,xcolor}
\usepackage[bottom]{footmisc}
\usepackage{mathtools}
\usepackage[breaklinks=true,colorlinks=true,citecolor=blue]{hyperref}

\spnewtheorem{rmk}{Remark}[section]{}{\itshape}

\allowdisplaybreaks

\begin{document}
\mainmatter

\title{Is MC Dropout Bayesian?}

\titlerunning{Is MC Dropout Bayesian?}

\author{Lo\"ic Le Folgoc\inst{1} 
\and
Vasileios Baltatzis\inst{2} \and
Sujal Desai\inst{1,3} \and
Anand Devaraj\inst{3} \and
Sam Ellis\inst{3} \and
Octavio E. Martinez Manzanera\inst{2} \and
Arjun Nair\inst{4} \and	
Huaqi Qiu\inst{1} \and
Julia Schnabel\inst{2} \and
Ben Glocker\inst{1}}

\authorrunning{L. Le Folgoc et al.}

\institute{BioMedIA, Imperial College London, United Kingdom \\
\and Biomedical Engineering and Imaging Sciences, King’s College London, UK \\
\and The Royal Brompton \& Harefield NHS Foundation Trust, London UK \\
\and Department of Radiology, University College London, UK \\
\email{l.le-folgoc@imperial.ac.uk}
}

\tocauthor{}

\maketitle

\newcommand{\bigcomment}[2]{
  \begin{center}
  \fbox{\begin{minipage}{\linewidth}{\bf #1:} {\rm #2}\end{minipage}}
  \end{center}
}

\newcommand{\R}{\mathbb{R}}
\newcommand{\bbC}{\mathbb{C}}
\newcommand{\bbH}{\mathbb{H}}
\newcommand{\bbi}{\boldsymbol{\mathrm{i}}}
\newcommand{\bbj}{\boldsymbol{\mathrm{j}}}
\newcommand{\bbk}{\boldsymbol{\mathrm{k}}}
\newcommand{\calB}{\mathcal{B}}
\newcommand{\calC}{\mathcal{C}}
\newcommand{\calD}{\mathcal{D}}
\newcommand{\calE}{\mathcal{E}}
\newcommand{\calF}{\mathcal{F}}
\newcommand{\calG}{\mathcal{G}}
\newcommand{\calH}{\mathcal{H}}
\newcommand{\calI}{\mathcal{I}}
\newcommand{\calL}{\mathcal{L}}
\newcommand{\calM}{\mathcal{M}}
\newcommand{\calN}{\mathcal{N}}
\newcommand{\calO}{\mathcal{O}}
\newcommand{\calQ}{\mathcal{Q}}
\newcommand{\calR}{\mathcal{R}}
\newcommand{\calS}{\mathcal{S}}
\newcommand{\calT}{\mathcal{T}}
\newcommand{\calV}{\mathcal{V}}
\newcommand{\calX}{\mathcal{X}}
\newcommand{\calW}{\mathcal{W}}
\newcommand{\rmA}{\mathrm{A}}
\newcommand{\rmC}{\mathrm{C}}
\newcommand{\rmD}{\mathrm{D}}
\newcommand{\rmE}{\mathrm{E}}
\newcommand{\rmF}{\mathrm{F}}
\newcommand{\rmH}{\mathrm{H}}
\newcommand{\rmI}{\mathrm{I}}
\newcommand{\rmJ}{\mathrm{J}}
\newcommand{\rmL}{\mathrm{L}}
\newcommand{\rmM}{\mathrm{M}}
\newcommand{\rmQ}{\mathrm{Q}}
\newcommand{\rmR}{\mathrm{R}}
\newcommand{\rmU}{\mathrm{U}}
\newcommand{\rme}{\mathrm{e}}
\newcommand{\rmf}{\mathrm{f}}
\newcommand{\rmh}{\mathrm{h}}
\newcommand{\rmn}{\mathrm{n}}
\newcommand{\rmp}{\mathrm{p}}
\newcommand{\rmq}{\mathrm{q}}
\newcommand{\rmu}{\mathrm{u}}
\newcommand{\rmv}{\mathrm{v}}
\newcommand{\rmx}{\mathrm{x}}
\newcommand{\rmy}{\mathrm{y}}
\newcommand{\bmc}{\bm{c}}
\newcommand{\bme}{\bm{e}}
\newcommand{\bmf}{\bm{f}}
\newcommand{\bmh}{\bm{h}}
\newcommand{\bms}{\bm{s}}
\newcommand{\bmq}{\bm{q}}
\newcommand{\bmt}{\bm{t}}
\newcommand{\bmr}{\bm{r}}
\newcommand{\bmu}{\bm{u}}
\newcommand{\bmw}{\bm{w}}
\newcommand{\bmy}{\bm{y}}
\newcommand{\bmz}{\bm{z}}
\newcommand{\brmf}{\boldsymbol{\rmf}}
\newcommand{\brmh}{\boldsymbol{\rmh}}
\newcommand{\brmp}{\boldsymbol{\rmp}}
\newcommand{\brmx}{\boldsymbol{\rmx}}
\newcommand{\brmy}{\boldsymbol{\rmy}}
\newcommand{\brmA}{\boldsymbol{\rmA}}
\newcommand{\brmC}{\boldsymbol{\rmC}}
\newcommand{\brmD}{\boldsymbol{\rmD}}
\newcommand{\brmE}{\boldsymbol{\rmE}}
\newcommand{\brmH}{\boldsymbol{\rmH}}
\newcommand{\brmI}{\boldsymbol{\rmI}}
\newcommand{\brmL}{\boldsymbol{\rmL}}
\newcommand{\brmQ}{\boldsymbol{\rmQ}}
\newcommand{\brmR}{\boldsymbol{\rmR}}
\newcommand{\dis}{\displaystyle}
\newcommand{\T}{{\mkern-1.5mu\mathsf{T}}}
\newcommand{\Id}{\text{Id}}
\newcommand{\tr}{\text{tr}}
\newcommand{\etal}{{\it et al }}
\newcommand{\eg}{\textit{e.g.}\xspace}
\newcommand{\ie}{\textit{i.e.}\xspace}
\newcommand{\uline}[1]{\underline{#1}}
\newcommand{\duline}[1]{\underline{\underline{#1}}}
\newcommand{\eq}{\!=\!}
\newcommand{\mrho}{{\!\rho}}
\newcommand{\iid}{{\textit{i.i.d.}}}

\begin{abstract}
MC Dropout is a mainstream ``free lunch'' method in medical imaging for approximate Bayesian computations (ABC). 
Its appeal is to solve  
out-of-the-box the daunting task of ABC and uncertainty quantification in Neural Networks (NNs); 
to fall within the variational inference (VI) framework; 
and to propose a highly multimodal, faithful predictive posterior. 
We question the properties of MC Dropout for approximate inference, 
as in fact MC Dropout changes the Bayesian model; 
its predictive posterior assigns $0$ probability to the true model on closed-form benchmarks; 
the multimodality of its predictive posterior is not a property of the true predictive posterior but a design artefact. 
To address the need for VI on arbitrary models, we share a generic VI engine within the pytorch framework. 
The code includes a carefully designed implementation of structured (diagonal plus low-rank) multivariate normal variational families, and mixtures thereof. 
It is intended as a go-to no-free-lunch approach, addressing shortcomings of mean-field VI with an adjustable trade-off between expressivity and computational complexity.
\end{abstract}

\section{Introduction}

The Bayesian framework provides a formalism for probabilistic predictions given partial knowledge of a potentially biased model of the real-world, and limited observations. Uncertainty quantification can improve risk assessment and decision making. It is especially relevant in medical imaging, given the complex relationship between the low-level processing (often performed sequentially) and the downstream, high-level patient management. Applications span segmentation~\cite{patenaude2011bayesian,iglesias2013improved}, registration~\cite{risholm2013bayesian,heinrich2016deformable,folgoc2017quantifying,schultz2018multilevel} and model personalisation~\cite{konukoglu2011efficient,mirams2016uncertainty,dhamala2018quantifying}. Crucial to Bayesian UQ is Bayesian modelling: ultimately the probabilistic model has to be faithful to the phenomenon it describes. Even moderately complex models bring forth computational challenges, hence second to faithful modelling is faithful approximate Bayesian computation (ABC). ABC engines aim at approximating intractable posterior distributions. Our main message is a warning against the misuse of MC dropout for ABC. We show that the MC dropout approximate posterior poorly fits the original model and is essentially non-Bayesian.

\begin{figure}[t]
\includegraphics[width=\textwidth]{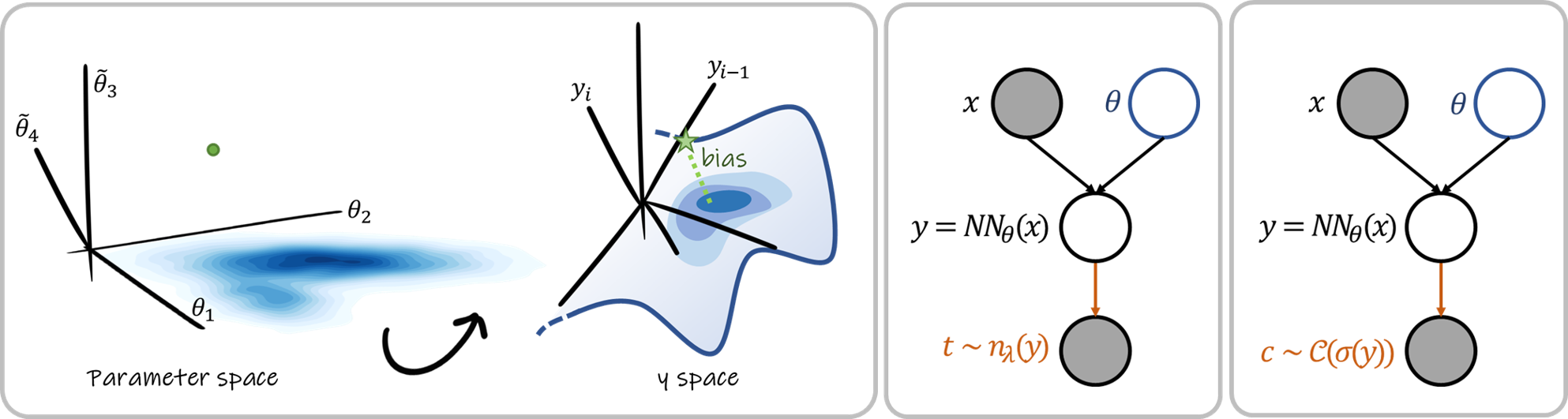}
\caption{(Left) The predictive error is due to a combination of model discrepancy/bias and within-model uncertainty. E.g. parameters $\tilde{\theta}_3,\tilde{\theta}_4$ fail to be accounted for by a $2$-parameter model. The true model (green dot) is not reachable by the parametrization $\theta_1,\theta_2$. Reachable solutions lie on a $2$D manifold (green star $\equiv$ true value). The predictive posterior on $y$ (blue shades, $y$-space) reflects within-model uncertainty, \ie uncertainty on the model parameters captured in the posterior (blue shades, $\theta$-space). (Middle) Standard Bayesian model of regression, resp. (Right) classification. Within-model uncertainty has epistemic ($\equiv$ blue node) and aleatoric ($\equiv$ orange arrow) sources.} \label{fig: Bayesian modelling}
\end{figure}

The predictive error (Fig.~\ref{fig: Bayesian modelling} left) is the combination of an out-of-model component, the model \textit{discrepancy} or \textit{bias}~\cite{kennedy2001bayesian} w.r.t. the real-world; and a within-model uncertainty whose nature is partly \textit{aleatoric} (due to noisy observations), partly \textit{epistemic} (due to unknown model parameters). Consider the regression model of Fig.~\ref{fig: Bayesian modelling} (middle), whereby a variable of interest $y$ results from some process $\text{NN}_{\theta}(x)$ that depends on inputs $x$ and unknown parameters $\theta$. Given this model choice and a dataset of noisy observations $\{X,T\}$, the knowledge of $y_\ast$ for new inputs $x_\ast$ is optimally described\footnote{in the sense of Bayesian risk minimization} by the predictive posterior distribution:
\begin{equation}
p(y_\ast| x_\ast, X, T) = \int_\theta p(y_\ast|x_\ast,\theta) p(\theta|X,T) d\theta\, , 
\label{eq: predictive posterior}
\end{equation}
which weighs the likelihood $p(y_\ast|x_\ast,\theta)$ by the posterior probability $p(\theta|X,T)$ and sums over all possible values of $\theta\!\in\!\Omega$ in the parameter space. In general Eq.~\eqref{eq: predictive posterior} has no closed-form and gives rise to a combinatorial problem. Several approximation strategies have been proposed, including Maximum A Posteriori inference, Markov Chain Monte Carlo~\cite{chen2016bridging,chen2018unified,gong2018meta}, Expectation Propagation~\cite{minka2013expectation,sun2017learning} and Variational Inference~\cite{zhang2018advances}. A common approach is to draw samples $\theta_k$ from the posterior $p(\theta|X,T)$ or from an approximation $q(\theta)$, followed by Monte Carlo integration, yielding:
\begin{equation}
p(y_\ast| x_\ast, X, T) \simeq \frac{1}{K}\sum_{1\leq k\leq K}p(y_\ast|x_\ast,\theta_k) \tag{\ref*{eq: predictive posterior}${}^\prime$} \, .
\label{eq: MC integration}
\end{equation}
When using an approximate posterior $q(\theta)$, the approximating family conditions the quality of the approximation. The family should be easy to sample from, rich enough to closely match the true posterior without making it overly challenging for optimizers to find a good fit $q^\ast$. VI approaches (incl. MC dropout~\cite{gal2016dropout}) can be analysed in light of the corresponding choice of $q$ (sections \ref{sec: background},\ref{sec: answer}). 
We compare MC dropout to alternative variational approximations: MAP, mean-field (MF-VI, with diagonal-covariance normal distributions), and finally structured normal distributions (sN-VI) or mixtures thereof (sGMM-VI). We contribute the variational engine for the latter (section \ref{sec: method}).
\section{Variational Inference And MC Dropout}
\label{sec: background}

VI aims to obtain a distribution $q(\theta)\!\in\!\calQ$ that best fits the true posterior among the chosen variational family $\calQ$, so as to exploit predictive estimates like Eq.~\eqref{eq: MC integration}. VI proceeds by maximizing the Evidence Lower-BOund (ELBO):
\begin{align}
\calL(q) & \triangleq \log{p(X,T)} - \text{KL}\left[q(\theta)\Vert p(\theta|X,T)\right] \, , \label{eq: ELBO} \\
\, & = \left\langle \log{p(T|X,\theta)} + \log{p(\theta)} \right\rangle_q + H(q) + \text{cst.}\, , \tag{\ref*{eq: ELBO}${}^\prime$} \label{eq: ELBO optimization}
\end{align}
where $H(q)=-\langle\log{q}\rangle_q$ is the entropy of $q$. Eq.~\eqref{eq: ELBO} establishes the equivalence between maximizing the ELBO and minimizing the (positive) Kullbach-Leibler divergence between true and variational posteriors.  
Eq.~\eqref{eq: ELBO optimization} makes the connection with standard penalized optimization clear. Observations $\{x_n,t_n\}$, $n\eq 1\cdots N$, are often assumed \iid, so that the gradient of $\log{p(T|X,\theta)}\eq \sum_n \log{p(t_n|x_n,\theta)}$ splits into individual sample contributions. Thus SGD and variants, using unbiased mini-batch gradient estimates, are suitable optimizers for the ELBO. By specifying the variational family $\calQ$ we retrieve various approaches.\\

\noindent
\textbf{MAP.} $q(\theta)\!\triangleq\!\delta_{\hat{\theta}}(\theta)$ for some $\hat{\theta}\!\in\!\Omega$. $q$ 
places all the probability mass at $\hat{\theta}$. The expectation in Eq.~\eqref{eq: ELBO optimization} collapses into the point evaluation at $\hat{\theta}$. $H(q)$ is a constant ($-\infty$). The global optima for $\hat{\theta}$ are the global mode(s) of $p(\theta|X,T)$.\\

\noindent
\textbf{Mean Field.} $q$ is non-parametric but $q(\theta)\!\triangleq\!\prod_i q_i(\theta_i)$ factorizes over parameters $\theta_i$, with generalizations to groupwise factorizations. The local extrema satisfy a set of coupled equations $\log{q_i^*(\theta_i)}\eq \langle f(X,T,\theta)\rangle_{q_j^*,j\!\neq\!i}$ that are closed form for hierarchical conjugate exponential models. This suggests iterative optimization of individual factors as in VBEM~\cite{bishop2006pattern}. Alternatively $q_i$ can be chosen among parametric families, leading to a computationally convenient subcase of what follows. \\

\noindent
\textbf{Parametric VI.} $q(\theta)\!\triangleq\!q_\psi(\theta)$ is a parametric family indexed by $\psi$ e.g., for multivariate normal distributions $\psi\eq(\mu,\Upsigma)$ are the mean and covariance matrix. Optimization is done on $\psi^\ast$. The \textit{Bayes by Backprop} strategy~\cite{blundell2015weight} combines stochastic gradient backpropagation with the \textit{reparametrization} trick. The trick uses an equivalent functional form for random draws $\theta\eq f(z,\psi)$ from $q_\psi$, using draws $z$ from a parameter-free distribution and an a.-e. differentiable $f$. E.g. $\theta\eq\mu\!+\!\rmL z$ with $\Upsigma\eq\rmL \rmL^\T$ and $z\sim\calN(0,\rmI)$ in the Gaussian case. An unbiased estimate of Eq.~\eqref{eq: ELBO optimization} is formed by Monte Carlo integration, replacing the expectation with an empirical average built from draws $\theta_k$; and backpropagated end-to-end onto the variational parameters $\psi$. Applied to $P$-dimensional NN parameters, the strategy is subject to the curse of dimensionality as 
$\psi$ can have a large memory footprint e.g., a full-rank covariance matrix has $\calO(P^2)$ parameters; and 
the variance of the stochastic gradients can cause convergence issues, calling for more samples and model evaluations. This creates an effective trade-off between simplicity and richness of the variational family.\\

\noindent
\textbf{Implicit distributions.} The approach uses the functional form $\theta\!\coloneqq\! f(z;\psi)\!\triangleq\!f_\psi(z)$ with $z\sim p_z$ a random draw from a standard parameter-free distribution, to implicitely define the variational distribution $q(\theta)$. The flexibility in the mapping $f_\psi$ (say, using NNs) allows to capture and sample from complex distributions more faithful to the true posterior. Unfortunately, $H(q)$ in Eq.~\eqref{eq: ELBO optimization} involves the log-density $\log{q(f_\psi(z))}$ of the \textit{pushforward} distribution $q(\theta)$ of $p_z$ by the map $f_\psi$, and is non-trivial to evaluate, giving rise to dedicated strategies~\cite{louizos2017multiplicative,huszar2017variational,shi2017kernel,tran2017deep}. These techniques have unparalleled  expressivity but currently involve sophisticated training strategies or limiting constraints such as invertibility of $f_\psi$. \\

\noindent
\textbf{MC dropout.} $q(\theta)$ is more easily described in an algorithmic way, as a sampling procedure for Monte Carlo integration of Eq.~\eqref{eq: ELBO optimization}. A binary variable $z_i\sim\calB(p)$ and a weight $\hat{\theta}_i$ are attached to every NN parameter $\theta_i$. At each iteration $z_i$'s are sampled, and $\theta_i\!\coloneqq\! \hat{\theta}_i z_i$ is forced down to $0$ if $z_i\eq 0$, set to $\hat{\theta}_i$ otherwise. $\hat{\theta}$ is optimized by stochastic backpropagation. The process mimics the training of dropout architectures~\cite{srivastava2014dropout}. Sampling proceeds similarly at test time to get an MC estimate Eq.~\eqref{eq: MC integration}. Mathematically there are $2^P$ joint states for $z\!\triangleq\!(z_i)_i$, with probabilities $\triangleq\! q(z)$ modulated by the activation probability $p$, yielding a mixture of delta-Dirac variational posterior parametrized by $\hat{\theta}$:
\begin{equation}
q(\theta) = \sum_{1\leq i\leq P} \sum_{z_i=0,1} 
\underbrace{
\delta_{(\hat{\theta}_1 z_1,\cdots, \hat{\theta}_P z_P)}(\theta_1,\cdots,\theta_P)
}_{\text{point mass at }\hat{\theta}\,\odot\, z} \,\cdot\, 
\underbrace{
\vphantom{ \delta_{(\hat{\theta}_1 z_1,\cdots, \hat{\theta}_p z_p)} }
p^{\sum z_i} (1-p)^{\sum 1-z_i}
}_{\text{$q(z)$ of joint state $z$}}\, ,
\label{eq: MC dropout posterior}
\end{equation}
where $\odot$ stands for elementwise multiplication. The MC dropout predictive posterior, with $q(\theta)$ replacing $p(\theta|X,T)$ in Eq.~\eqref{eq: predictive posterior}, follows in closed form as:
\begin{equation}
p_{\text{MC-dropout}}(y_\ast| x_\ast, X, T) = \sum_{z\in [0,1]^P} 
p(y_\ast|x_\ast,\hat{\theta}\!\odot\! z) \, \cdot q(z)
\label{eq: MC dropout predictive}
\end{equation}

\section{Is MC Dropout Bayesian?}
\label{sec: answer}

\begin{figure}[t]
\includegraphics[width=\textwidth]{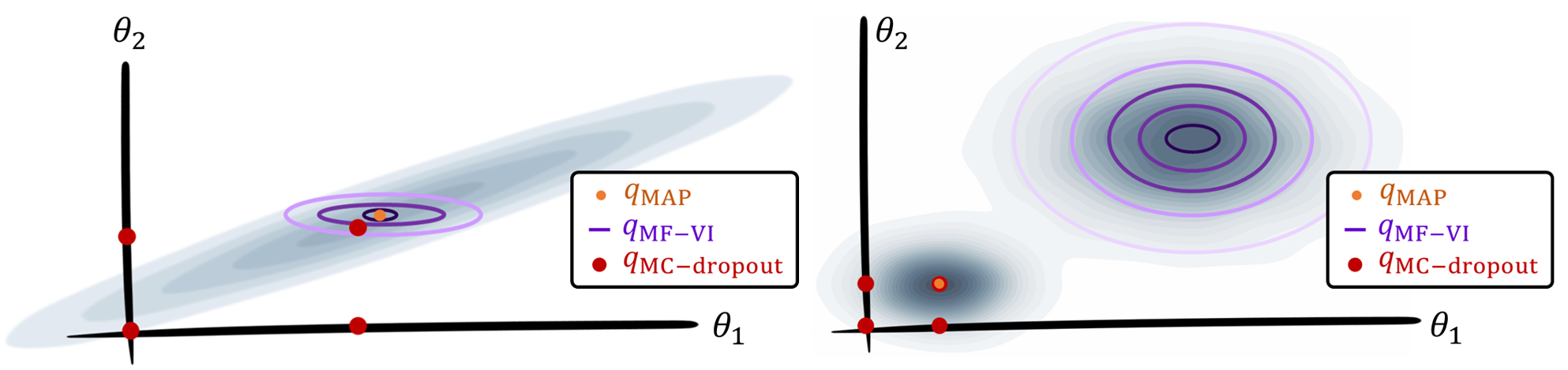}
\caption{$2$D illustration of variational posterior fits, when the true posterior is unimodal (Left) or bimodal (Right). The true posterior is displayed as a blue-grey probability heatmap. Delta-Dirac masses are displayed as dots (MC dropout in red, MAP in orange). The MAP $\delta$-Dirac is at the posterior mode. MC dropout is a mixture of $2^2\eq 4$ $\delta$-Dirac masses. The Gaussian MF-VI fit is displayed as axis-aligned isocontours.} \label{fig: posterior}
\end{figure}

As shown in Fig.~\ref{fig: posterior}, MC dropout approximations fail sanity checks in a similar way to MAP approximations, in that they only assign non-zero probability to a finite set of parameters $\theta$. In addition it is a modal approximation, hence sensitive to points of high posterior rather than to regions of high probability mass (Fig.~\ref{fig: posterior}, right). The resulting predictive posterior is illustrated in Fig.~\ref{fig: MC dropout samples} in toy NN regression examples. All of the probability mass is distributed on a finite set of models (Eq.~\eqref{eq: MC dropout predictive}), and the true model is assigned $0$ probability. This contrasts with all other choices of variational families.

Mean-field (MF-VI) approximations have non-degenerate probability density functions but cannot capture covariance between parameters, hence are known to suffer from uncertainty under-estimation. This is best seen from Fig.~\ref{fig: posterior} where the Gaussian MF-VI posterior is restricted to axis-aligned isolines (ellipsoids). Full-covariance normal distributions have the expressivity to capture parameter covariance. Structured-covariance normal VI (sN-VI) lie within the two extremes, with $\Upsigma = \text{diag}(\rmA) + \rmU \rmU^\T$ where $\rmU\in\R^{P\times K}$ is low-rank ($K\ll P$). These models are unimodal approximations and fit complex multimodal posteriors poorly. Mixture models address this limitation.

MC dropout always yields a multimodal posterior (Eq.~\eqref{eq: MC dropout posterior}\eqref{eq: MC dropout predictive}), even when the true posterior is unimodal. Despite being multimodal, it only has a single $P$-dimensional degree of freedom $\hat{\theta}$ that  
decides the mass distribution: 
delta-Dirac point-masses are located at $\hat{\theta}$ and at its (orthogonal) projections on every $1$D axis, $2$D plane, ..., $(P-1)$D hyperplane through the origin. This is a design artefact unrelated to the properties of the Bayesian model or of the data. In particular MC dropout does not have the expressivity to capture multimodal posteriors with its single degree of freedom $\hat{\theta}$ (Fig.~\ref{fig: posterior}, right). Moreover as the strategy would clearly be ill-suited for NN bias parameters, for which MC dropout therefore reverts to a standard $\delta$-Dirac pointwise estimate.\\

\noindent
\textbf{Alternative interpretation.} The MC dropout posterior is controlled by an arbitrary probability $p$ of neuron activation. When $p\!\coloneqq\! 1$, the posterior coincides with a MAP approximation $q_{\text{MAP}}$, whereas $p\!\coloneqq\! 0$ places all the mass at the origin $\hat{\theta}\eq0$. In practice the user manually chooses a value that optimizes say, accuracy. Rather than a technique for ABC, we can interpret the MC dropout strategy as modifying the original Bayesian model with a sparsity-inducing prior. Each parameter $\theta_i$ is attached a Bernoulli variable $z_i\!\sim\!\calB(p)$. Then MC dropout follows as a naive approximation $q(\theta,z)=\delta_{\hat{\theta}}(\theta)p_z(z)$, whereby a (non-Bayesian) modal approximation is used on $\theta$ and $z$ is sampled according to the prior $p_z$ (instead of fitting say, $q(z)=\prod_i \calB(z_i;p_i)$ or a logistic normal).  

\section{No Free Lunch Variational Inference}
\label{sec: method}

We contribute an engine for parametric VI based on the reparametrization trick and stochastic backpropagation\footnote{The code is available at \url{https://github.com/llefolgo/dlvi}}, with an implementation of structured multivariate normal families (sN-VI) and mixtures thereof (sGMM-VI). These are low-parametric families that capture parameter covariance within and across layers (and sGMM is multimodal). The implementation uses the pytorch framework, and reuses the design of the tool \textit{pyvarinf}~\cite{pyvarinf} to ``variationalize'' an arbitrary input model (e.g. NN). \textit{Pyvarinf} implements Gaussian MF-VI. It rebuilds and evaluates the model on the fly (on a minibatch) with the sampled parameters $\theta$ after drawing from the variational posterior $q_{\psi,i}(\theta_i)=\calN(\theta_i;\mu_i,\sigma_i)$, so that the model is seamlessly autodifferentiable w.r.t. $\psi$. We extend the mechanism to use arbitrary variational families $q_\psi$, given a functional rule $\theta=f(z,\psi)$ to draw one or multiple samples; and an evaluation of the log-density $\log{q_\psi(\theta)}$ or of the entropy $H(q_\psi)$. To initialize variational parameters, the default behaviour exploits heuristics based on the weight initialization routines of the original pytorch layers. As a training objective, one typically combines the MC estimate of the ELBO in Eq.~\eqref{eq: ELBO optimization} with any out-of-the-box stochastic gradient optimizer. The prior $p(\theta)$ is a Bayesian modelling choice and arbitrary. We provide examples using $l2$-regularization, (quasi) scale-invariant Student-t distributions and/or end-to-end Lipschitz regularization of the feature maps.

sN-VI and sGMM-VI have been proposed by Miller et al.~\cite{miller2017variational} in the context of iterative, incremental refinement of an existing variational posterior. The authors leverage computational gains via low-rank (Woodbury) matrix identities, which we exploit as well. We depart from the literature based on the observation that standard stochastic backpropagation exhibits convergence issues even on sanity checks and with simple low-parametric families like sN. 
To stabilize gradient updates, the implementation proposes \textit{paired} and \textit{unscented} modes whereby coupled samples are drawn at once, in place of the naive reparametrization trick. In the paired mode, twin samples symmetrized around the mean are drawn. The unscented mode draws $K$ coupled samples that use jointly orthogonal random combinations of the $K$ low-rank directions, and their twin symmetrics; somewhat reminiscent of unscented Kalman filtering. This disambiguates contributions of the mean, diagonal and low-rank covariance terms in the gradient updates. In addition we advocate as in~\cite{blundell2015weight} the use of $\log{q_\psi(\theta_k)}$ with the sampled $\theta_k$'s in the Monte Carlo integration of Eq.~\eqref{eq: ELBO optimization} rather than say, the closed-form entropy $H(q_\psi)$ even when it is available. It reduces the variance of stochastic updates by letting draw-dependent scaling effects affect equally all terms.  
\section{Examples}
\label{sec: results}

\noindent
\textbf{Gaussian distribution fit.} To validate the implementation of the variational engine, we fit a random multivariate $8$-dimensional normal distribution $p=\calN(\mu_0,\Sigma_0)$ via MAP, Gaussian MF-VI and various low-rank (from $1$ to $8$) structured normal distributions $\text{sN-VI}_{lr}$. Table \ref{table: Gaussian distribution fit} reports the Kullback-Leibler divergence $\text{KL}[p||q]$ from the true distribution $p$ to the variational posterior $q$. The pointwise MAP approximation yields an infinite divergence. MC-dropout is the only variational method not natively able to handle this sanity check (although a naive application of section \ref{sec: background} yields $\text{KL}[p||q]=+\infty$). As expected, the various $\text{sN-VI}_{lr}$ are natural in-between approximations between a diagonal approximation like MF-VI ($=\text{sN-VI}_0$) and a full-covariance approximation like $\text{sN-VI}_8$.

\begin{table}\centering
\small
\tabcolsep=0.14cm
\begin{tabular}{c *{7}c}
 \, & $\text{MAP}$ & $\text{MC-drop.}$ & \text{MF-VI} & $\text{sN-VI}_{1}$ & $\text{sN-VI}_{2}$ & $\text{sN-VI}_{4}$ & $\text{sN-VI}_{8}$ \\ \midrule
${\text{KL}[p||q]}$ & $+\infty$ & --- & $175.8051$ & $37.9890$& $25.9107$& $0.8774$& $0.0979$ \\ 
\end{tabular}
\vspace{0.1in}
\caption{Quality of fit ${\text{KL}[p||q]}$ of various variational approximations to a multivariate normal.}
\label{table: Gaussian distribution fit}
\end{table}

\noindent
\textbf{RBF regression.} To illustrate the mechanism behind MC-dropout stochasticity we fit a small RBF 1-layer model in absence of model bias. Namely, let $c_k$, $1\leq k\!\leq\! K\eq 10$, a set of regularly spaced basis centers. Let the true model $y = f(x) \triangleq \sum_{1\leq k\leq 10} K(c_k,x) \theta_k^\ast$ a radial basis function with the $\theta_k^\ast\sim \calN(0,1)$ sampled randomly from a Gaussian distribution. Observations $t_n= y_n + \epsilon_n$, $\epsilon_n\sim\calN(0,\sigma^2)$, $\sigma:=0.25$, are drawn with Gaussian noise, forming a training dataset $\{X,T\}=\{x_n,t_n\}_{n\leq N}$. Since the MC-dropout algorithm has no native mechanism to jointly fit the aleatoric noise, we assume the noise model known for simplicity. The full model is a one-layer linear model, $\mathbf{y}=\Uppi_{\mathbf{c}}(\mathbf{x})\mathbf{\uptheta}$, with $\mathbf{y}=(y_n)_{n\leq N}$, $\mathbf{\uptheta}=(\theta_k)_{k\leq K}$, and $\Uppi_{\mathbf{c}}(\mathbf{x})$ the matrix of $(n,k)$-th coefficient $K(c_k,x_n)$. We again fit the model with various methods, from MAP and MC-dropout to MF-VI and sN-VI. At training time, MC dropout fits a set $\hat{\mathbf{\uptheta}}$ of weights by randomly dropping some components to $0$ in $\mathbf{\uptheta}:=\hat{\mathbf{\uptheta}}\!\odot\! \mathbf{z}$ with $z_k$'s randomly set to $0$ or $1$. 

\begin{figure}
\centering
\includegraphics[width=0.75\textwidth]{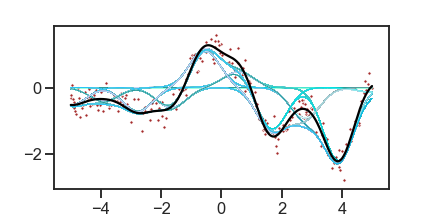}
\caption{MC-dropout posterior on 1-layer RBF regression. Black line: true model. Brown dots: observations. Blue lines are not samples from the approximate posterior, but the set of $1024$ delta-Dirac functions forming the multimodal MC-dropout posterior. Evidently the location of the modes is unrelated to the model uncertainty and is instead simply an artefact of the MC-dropout approximation.} \label{fig: MC dropout samples}
\end{figure}

Fig.~\ref{fig: MC dropout samples} summarizes the MC-dropout posterior as a set of $2^{10}=1024$ delta-Dirac functions, a.k.a.~"samples", generated by the $\mathbf{\uptheta}^{(s)}:=\hat{\mathbf{\uptheta}}\odot\mathbf{z}^{(s)}$, $1\leq s\leq 1024$, obtained from the various combinations of $0$ and $1$'s as per section~\ref{sec: background}. None of the delta Dirac functions coincide with the true model (\ie $\mathbf{\uptheta}^{(s)}\!\neq\!\mathbf{\uptheta}^\ast$ for all $s$). Thus the true model has $0$ probability under the MC-dropout approximation. Under Gaussian variational approximations $q(\mathbf{\uptheta})=\calN(\mathbf{\uptheta}|\upmu_v,\Upsigma_v)$ instead, considering $N_\ast>K$ (unseen) test points $\{X_\ast,Y_\ast\}$, the probability of the true model $\mathbf{y}_\ast \eq \Uppi_{\mathbf{c}}(\mathbf{x}_\ast)\mathbf{\uptheta}^\ast$ is $\calN(\mathbf{\uptheta}^\ast|\upmu_v,\Upsigma_v)=q(\mathbf{\uptheta}^\ast)\!\neq\! 0$.

Finally the exact posterior for $\mathbf{\uptheta}$ is closed form as $p^\ast\triangleq p(\mathbf{\uptheta}|X,T)=\calN(\upmu, \Upsigma)$, $\Upsigma\!\triangleq\! (\sigma^{-2}\Uppi_{\mathbf{c}}(X)^T\Uppi_{\mathbf{c}}(X) + \brmI)^{-1}$, $\upmu\eq\sigma^{-2}\Upsigma \Uppi_{\mathbf{c}}(X)^T\mathbf{t}$. The quality of approximation $\text{KL}[p^\ast||q]$ of the approximate posteriors $q(\mathbf{\uptheta})$ is also reported in Table~\ref{table: regression fit}. All Gaussian variational approximations perform similarly here.

\begin{table}\centering
\small
\tabcolsep=0.14cm
\begin{tabular}{c *{7}c}
 \, & $\text{MAP}$ & $\text{MC-drop.}$ & \text{MF-VI} & $\text{sN-VI}_{1}$ & $\text{sN-VI}_{2}$ & $\text{sN-VI}_{4}$ & $\text{sN-VI}_{10}$ \\ \midrule
$\log{q(\mathbf{\uptheta}^\ast)}$ & $-\infty$ & $-\infty$ & $15.5161$ & $14.9543$& $14.6339$& $13.8562$& $15.1211$ \\ 
${\text{KL}[p^\ast||q]}$ & $+\infty$ & $+\infty$ & $3.1853$ & $1.6967$& $3.8009$& $4.2452$& $0.8389$ \\ 
\end{tabular}
\vspace{0.1in}
\caption{Quality of fit of variational approximations on an RBF regression task.}
\label{table: regression fit}
\end{table}

\bibliographystyle{styles/bibtex/spmpsci}
\bibliography{bibliography}

\end{document}